\documentclass[accepted,startpage]{uai2022}
\usepackage[american]{babel}
\usepackage{natbib} % has a nice set of citation styles and commands
\usepackage{mathtools} % amsmath with fixes and additions
\usepackage{booktabs} % commands to create good-looking tables
\usepackage{tikz} % nice language for creating drawings and diagrams

\usepackage{graphicx}
\usepackage{amsmath}
\usepackage{amssymb}
\usepackage{booktabs}
\usepackage{amsfonts}
\usepackage{amsmath}
\usepackage{xcolor}
\usepackage{multirow}
\usepackage{multicol}
\usepackage{makecell}

\usepackage{caption}
\usepackage{subcaption}
\usepackage[capitalize]{cleveref}
\crefname{section}{Sec.}{Secs.}
\Crefname{section}{Section}{Sections}
\Crefname{table}{Table}{Tables}
\crefname{table}{Tab.}{Tabs.}

\newcommand{\BEAS}{\begin{eqnarray*}}
\newcommand{\EEAS}{\end{eqnarray*}}
\newcommand{\BEA}{\begin{eqnarray}}
\newcommand{\EEA}{\end{eqnarray}}
\newcommand{\BEQ}{\begin{equation}}
\newcommand{\EEQ}{\end{equation}}
\newcommand{\BIT}{\begin{itemize}}
\newcommand{\EIT}{\end{itemize}}
\newcommand{\BNUM}{\begin{enumerate}}
\newcommand{\ENUM}{\end{enumerate}}

\newcommand{\BA}{\begin{array}}
\newcommand{\EA}{\end{array}}

\newcommand{\reals}{{\mathbb R}}

\newcommand{\norm}[1]{\left\lVert#1\right\rVert}

\title{Complementing Semi-Supervised Learning with Uncertainty Quantification}

\author[1]{\href{mailto:<ehsan.kf@gmail.com>?Subject=Your UAI 2022 paper}{Ehsan~Kazemi}{}}
\affil[1]{%
    Computer Science Dept.\\
    University of Central Florida\\
    Orlando, Florida, USA
}
\begin{document}

\maketitle

\begin{abstract}
    The problem of fully supervised classification is that it requires a tremendous amount of annotated data, however, in many datasets a large portion of data is unlabeled. To alleviate this problem semi-supervised learning (SSL) leverages the knowledge of the classifier on the labeled domain and extrapolates it to the unlabeled domain which has a supposedly similar distribution as annotated data. Recent success on SSL methods crucially hinges on thresholded pseudo labeling and thereby consistency regularization for the unlabeled domain. However, the existing methods do not incorporate the uncertainty of the pseudo labels or unlabeled samples in the training process which are due to the noisy labels or out of distribution samples owing to strong augmentations. Inspired by the recent developments in SSL, our goal in this paper is to propose a novel unsupervised uncertainty-aware objective that relies on aleatoric and epistemic uncertainty quantification. Complementing the recent techniques in SSL with the proposed uncertainty-aware loss function our approach outperforms or is on par with the state-of-the-art over standard SSL benchmarks while being computationally lightweight. Our results outperform the state-of-the-art results on complex datasets such as CIFAR-100 and Mini-ImageNet.  
\end{abstract}

\section{Introduction}

\begin{figure}[t]
  \centering
    \subfloat{\includegraphics[width=0.42\linewidth]{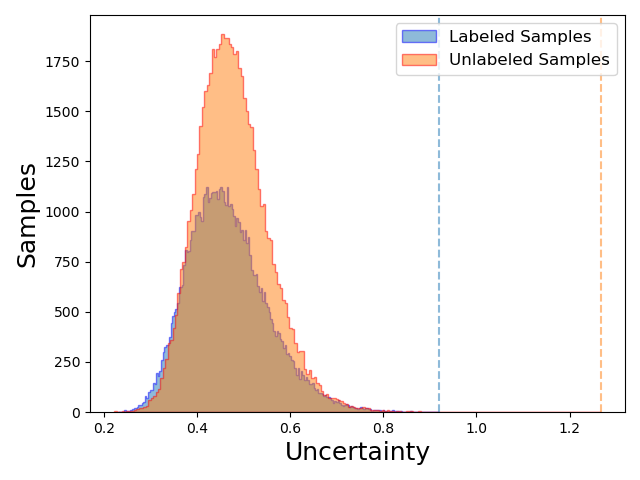}}\hfill
   \subfloat{\includegraphics[width=0.45\linewidth]{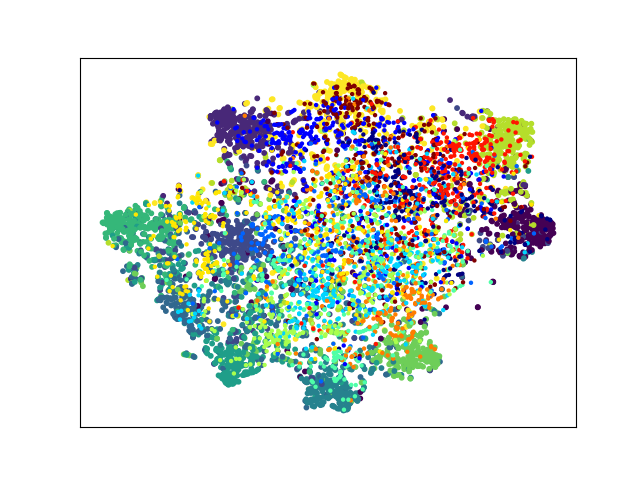}}\\
   \subfloat{\includegraphics[width=0.42\linewidth]{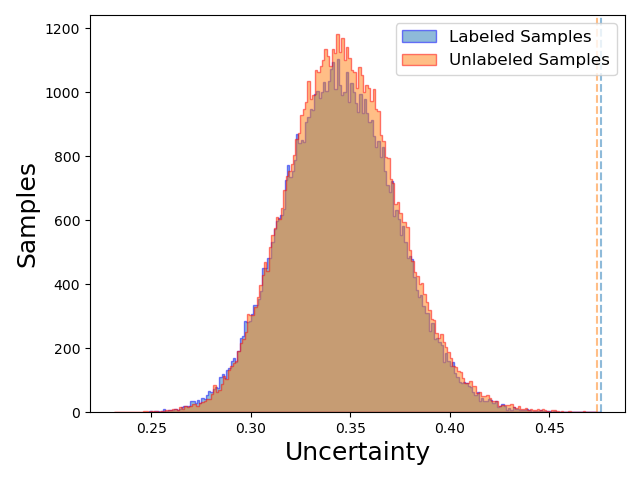}}\hfill
   \subfloat{\includegraphics[width=0.45\linewidth]{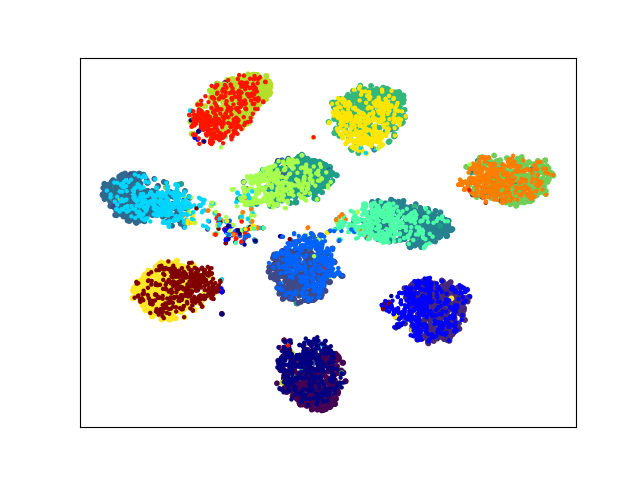}}
   \caption{ Estimation of epistemic uncertainty (a) for a model trained in a fully supervised manner on 4000 labeled examples of CIFAR-10 dataset; and (c) for a model trained using our SSL approach over CIFAR-10 with 4000 labeled samples. The 1 quantile of each distribution is shown by a dashed line. The joint t-SNE of the corresponding (b) supervised model; and (d) SSL model, for the embedding of weakly augmented labeled training samples (dimmer color) and strongly augmented unannotated training samples (brighter color). The figures show that our proposed model clusters unlabeled domains consistently with the clusters of labeled data. The distributions of epistemic certainty of labeled and unlabeled samples in the histogram (c) are similar, while the epistemic uncertainty distributions for the supervised model for labeled and unlabeled samples in (a) are significantly distinguished.}
   \label{fig-epistemic-uncertainty-tsne}
\end{figure}

Deep learning has seen unprecedented success by providing a remarkable performance on tasks such as computer vision, natural language processing, and virtual reality. One of the biggest challenges for deep models is the lack of annotated data, while the amount of annotated data for training is the major factor in the performance of deep models. Therefore, the performance boost gained by the use of large datasets is at a cost for data annotation. However, the large labeled dataset is not always available due to the difficulty or infeasibility of data annotation mechanisms. Further, annotating the labels  in many domains is either expensive or infeasible because of accessibility and privacy constraints. Semi-supervised learning (SSL) is a paradigm to fill this void. Semi-supervised training is becoming significantly important as abundant unlabeled data is used to improve the generalization performance of deep models. The ultimate goal of SSL is to have better performance in scarce-label regimes as it eases the requirement of annotated data. Since unlabeled data can be obtained at minimal cost, any performance benefit from SSL is attractive. Therefore, recently deep learning
has seen an excess of SSL methods  \citet{miyato2018virtual,sajjadi2016regularization,laine2016temporal,tarvainen2017mean,sajjadi2016mutual,verma2021interpolation,kuo2020featmatch,sohn2020fixmatch,berthelot2019remixmatch,hu2021simple}. SSL leverages the information on the labeled domain to use the unlabeled data in the training process. The main part of SSL is the way the information from the labeled domain is extrapolated to the unlabeled domain. The basic techniques in SSL employ the continuity assumption by taking into account that the data which are close in a proper metric would probably share the same labels. Following this path, many approaches in SSL have recently been developed \citet{jaakkola2002partially,zhou2004learning,douze2018low,chapelle2009semi,iscen2019label}. The introduced approaches for SSL mainly hinge on consistency regularization and pseudo labeling on the unannotated domain. Pseudo labeling is a mechanism for labeling the unannotated data by the model itself. These pseudo labels are used to supervise training of the model. A common practice for pseudo labeling is to apply a confidence threshold to select only the unlabeled samples with labels of high probability predictions \citet{lee2013pseudo,french2017self,sohn2020fixmatch}. The confidence in the context of SSL refers to the highest probability across the class predictions. By applying the confidence threshold for the pseudo labels we abstain from labeling the samples that are close to the decision boundaries. The core of recent SSL methods relies on unsupervised losses to train on both the labeled and unlabeled datasets annotated with artificial labels. Similarly, consistency regularization generates artificial labels on randomly modified inputs or models \citet{laine2016temporal,sajjadi2016regularization}. Recently developed SSL techniques exploit consistency regularization to learn from a large amount of data by contrasting the augmented unlabeled data and quantifying the consistency among the unlabeled samples \citet{hu2021simple,kuo2020featmatch}. The results for image classification using SSL based on consistency regularization have shown a front edge performance \citet{berthelot2019mixmatch,berthelot2019remixmatch,sohn2020fixmatch,hu2021simple}.

Traditionally, SSL methods do not account for the uncertainty for the pseudo labeling and assume that the model would not discriminate between labeled and unlabeled domains. Incorporating the uncertainties in SSL is crucial as pointed out in \citet{iscen2019label,tagasovska2019single}, since it abstains the sample to contribute to the loss in the highly uncertain regimes. We believe that it is very important for the model to quantify uncertainty when facing unlabeled data and to let the uncertainty contribute to the unsupervised loss formulation. Accurate quantification of uncertainties from unlabeled data and pseudo labels allow for the full potential of unlabeled data in SSL. In particular, when the model tries to extrapolate its knowledge to the unlabeled domain it is important to quantify and minimize the uncertainty to ensure robustness in distribution learning. The uncertainty attribution to the loss compensates for the weaknesses of the pseudo labeling mechanism and lets the model learn a distribution over the entire dataset without discriminating between labeled and unlabeled datasets.

 In this paper, we propose a method to leverage uncertainty to improve the performance of SSL methods. For this purpose, we mainly focus on heteroscedastic uncertainty which stands for estimating the varying uncertainty for each sample. We employ the existing approaches to quantify heteroscedastic and multivariate uncertainties for the SSL problem in order to improve their overall performance. Heteroscedastic uncertainties in deep learning have two main sources. The first one is the epistemic uncertainty or model uncertainty which is associated with the uncertainty of model parameters. Measuring epistemic uncertainty is an intractable problem and typically Bayesian inference is applied to estimate this type of uncertainty \citet{kendall2017uncertainties}. Accurate quantification of epistemic uncertainty results in more reliable performance of the machine learning models in out of distribution domains. The second one aleatoric uncertainty is related to the stochasticity in the distribution of data and is not reducible by increasing the model capacity or the training data. Accurate quantification of aleatoric uncertainty would maximize the performance of the model when fusing deep model prediction with conventional models by employing for example Kalman filter method. A practical example of aleatoric uncertainty in SSL is generating pseudo labels using the neural model itself which are noisy by construction. As in \citet{kendall2017uncertainties}, in this work, we decouple the uncertainties of aleatoric and epistemic sources and neglect the cross correlations between these two types of uncertainties. The assumption of independence of aleatoric and epistemic uncertainties is valid when epistemic uncertainty is small. 

In this work, we leverage the mechanisms in literature to quantify the uncertainty inherited from the unlabeled domain and pseudo labels and incorporate it into the loss function. The uncertainty quantification is performed in an end-to-end approach by harnessing the Gaussian likelihood and certificate regression for estimating the deep epistemic uncertainty \citet{tagasovska2019single}. Our uncertainty estimation is only a one-shot forward pass which is computationally efficient. Employing uncertainty quantification techniques, we develop an uncertainty-aware unsupervised loss that minimizes the aleatoric and epistemic uncertainty of both the distribution of unlabeled data and the knowledge extrapolated to the unlabeled domain to extract pseudo labels. In our proposed method the epistemic uncertainty presumably accounts for uncertainty due to knowledge extrapolation to low-density areas, while aleatoric uncertainty represents the uncertainty and noise owing to pseudo label generation and out of distribution unannotated examples.
The distribution of epistemic uncertainty of two models which are trained over 4000 labeled data of CIFAR-10 dataset is depicted in Figure \ref{fig-epistemic-uncertainty-tsne}. The figure shows the direct correlation of aligning the uncertainty distribution for labeled and unlabeled data to the clustering performance of the model. As it is observed from the figures when the uncertainty distribution of annotated data is distinguished from the uncertainty distribution of unlabeled samples, the model performs poorly on the unannotated domain. By minimizing the epistemic uncertainty we implicitly align the distribution of the model over labeled and unlabeled domains while training progresses. Minimizing the epistemic uncertainty pushes the decision boundaries of clusters to the low-density areas, which is also recommended in the low-density separation assumption \citet{chapelle2009semi}.

To precisely integrate our approach to SSL, the proposed algorithm leverages the state-of-the-art SSL works in MixMatch family \citet{berthelot2019mixmatch,berthelot2019remixmatch,sohn2020fixmatch} for thresholded pseudo labeling and consistency regularization for unannotated examples. Since we build our framework on the existing methods for SSL, it bears substantial similarities to the contemporary SSL algorithms. Following consistency regularization paradigm \citet{berthelot2019mixmatch,berthelot2019remixmatch,sohn2020fixmatch} our approach uses thresholded pseudo labeling by exploiting the model's prediction on weakly augmented unannotated images and thereby applying consistency regularization on strongly augmented examples. The proposed method does not use a pre-trained model to infer labels and the pseudo-labels are generated as the training progresses. Therefore, in our framework the uncertainty quantification of pseudo labels is crucial. The overall architecture of our method is shown in Figure \ref{fig-overall_arch}. We let the deep model assess its uncertainties when unlabeled data is presented to the model. In particular, the neural model is trained to output the aleatoric uncertainty and the certificate for approximation of epistemic uncertainty. Afterward, the unsupervised loss function is optimized based on uncertainty estimations from the model.  We develop a method that is simple but more accurate or on par with the state-of-the-art methods. Our approach is complementary to the current approaches in semi-supervised learning. In the experiments, we demonstrate the performance of the proposed uncertainty-aware algorithm in complementing the state-of-the-art SSL techniques.  Our framework is computationally attractive as it does not demand to form label propagation graphs and is free of computing similarities across the unlabeled examples as suggested in \citet{hu2021simple,iscen2019label}, while providing competitive results. To the best of our knowledge, this work is the first effort to estimate and align the uncertainty distribution of labeled and unlabeled data. The previous efforts in SSL were mainly devoted to enhancing the consistency across and within the strongly and weakly augmented unlabeled samples. 

Our contribution is summarized as follows:
1- We propose a novel uncertainty-aware objective that incorporates both the aleatoric and epistemic uncertainty to the SSL to align the uncertainty distributions of labeled and unlabeled domains. 
2- We implement our approach by complementing the contemporary work with the proposed uncertainty-aware framework. 
3- We evaluate our method across a variety of standard SSL benchmarks. We show that our method achieves state-of-the-art performance on standard SSL benchmarks. For example, our method obtains 79.32\% accuracy on CIFAR-100 
dataset with 10000 labeled samples compared to the state-of-the-art of 78.11\% \citet{hu2021simple}.
\section{Related Works}
The requirement of huge annotated data for learning deep models hampered their application in different domains. Although visual data is abundant, the data which is reliably annotated is scarce. Nevertheless annotating unlabeled data is not practical, expensive, and prone to errors and noisy labels. The literature on SSL is wealthy and different and distinct approaches to this topic were recently developed. In this review, we focus mainly on the works which employ Augmentation Anchoring and Consistency Regularization that bears the most resemblance to our implementation of the baseline SSL algorithm upon which we build our framework. Classical works in SSL are focused on transductive learning. 
In \citet{zhou2004learning,zhu2003semi} transductive learning by diffusion is applied to use the entire dataset and annotated data to infer labels for the unlabeled examples. Applying a classifier trained on annotated data can produce high-quality pseudo labels, which is providing a form of supervision for classifier training. The pseudo label data is used to augment the training data and the model is retrained until convergence. In \citet{iscen2019label} a label propagation method based on transductive label propagation is offered where entropy is used as a measure of uncertainty being incorporated into the diffusion matrix for pseudo labeling. In SSL the fully supervised performance is providing the upper bound. 

Thresholded pseudo labeling \citet{rosenberg2005semi} refers to a distinct version of SSL where confidence-based thresholding applies to model predictions to retain labels for unannotated examples only when the model is sufficiently  confident. Pseudo labeling has recently seen notable success as a part of SSL algorithms \citet{arazo2020pseudo}. Using the soft pseudo labels to train the unlabeled data, pseudo labeling can be seen as an entropy minimization \citet{grandvalet2005semi} which has been efficiently employed in SSL methods \citet{miyato2018virtual}. Random augmentations such as data augmentation \citet{french2017self}, stochastic regularization (e.g. Dropout \citet{srivastava2014dropout}) \citet{sajjadi2016regularization,laine2016temporal}, and adversarial perturbations \citet{miyato2018virtual} are employed in the pseudo labeling method.
In \citet{sajjadi2016mutual,laine2016temporal,tarvainen2017mean,qiao2018deep,miyato2018virtual} a so-called consistency loss is applied to both labeled and unlabeled examples and the consistency under the different transformations of samples or models is enhanced by minimizing the discrepancies in a proper metric. These methods are based on the fact that the model response to input should remain the same under the transformations that result in small perturbation in a proper metric. Consistency regularization was first introduced in \citet{bachman2014learning} and early versions of it use the  exponential moving average of model parameters \citet{tarvainen2017mean} and previous model checkpoint for generating pseudo labels \citet{laine2016temporal}. The recent results \citet{sohn2020fixmatch} show that strong augmentations in consistency matching provide better classification accuracy. In \citet{hu2021simple} and \citet{kuo2020featmatch} the consistency across the unlabeled samples is leveraged to improve the consistency regularization based SSL methods.

Uncertainty estimation and incorporating only highly confident predictions are very crucial in SSL. Neglecting the uncertainty in pseudo labeling may result in drifting from optimal parameters on noisy pseudo labels. The thresholded pseudo labels by confidence were used previously in sample selections and label propagation  \citet{lee2013pseudo,hu2021simple,iscen2019label,sohn2020fixmatch,iscen2019label}. Despite the importance of uncertainty assessment in machine learning models, the uncertainty quantification for deep models is still an open question, where the prior efforts are mainly attributed to Bayesian methods \citet{hernandez2015probabilistic,blundell2015weight,gal2016dropout,kendall2017uncertainties,khan2018fast,teye2018bayesian}, being computationally an intractable approach.
Several sample-based approaches have been developed to estimate epistemic uncertainty using Bayesian inference. The most practical approach is to apply dropout Monte Carlo \citet{gal2016dropout}, which is computationally expensive. Other approaches include Bayesian ensemble for uncertainty estimation \citet{lakshminarayanan2017simple}. In this work, we exploit the uncertainty of the model over unlabeled data to efficiently learn the actual distribution over the entire dataset constituting from labeled and unlabeled data. Our approach for quantifying uncertainty does not require the few-shot mechanism with an ensemble of models and therefore computationally it is efficient.

\section{Preliminaries}
In this section, we discuss the classifier, the pseudo labeling mechanism, and the supervised loss for pseudo labels that our method is based on. We consider a set of $n$ samples $X := \{x_i\}_{i=1}^n$, of a $h$ class set $\mathcal{X}$ with $L$ annotated samples denoted by $X_L$ with labels $Y_L = \{y_i\}_{i=1}^L$. The remaining of $u := n-L$ samples are unlabeled set denoted by $X_U := \{x_i\}_{i=L+1}^n$. Our goal in SSL is to leverage the entire set $X$ to learn a model which maps unseen samples to the labels. We let $f_\theta: \mathcal{X} \to \reals^h$, where $\theta$ is the model parameter. We divide the network for classification into two parts. The first part of the model is the feature extractor network $\phi_\theta: \mathcal{X} \to \reals^d$, projecting the input sample to the feature vector. The second part maps the feature vector to the confidence score. This part typically consists of a block of fully connected layers. Therefore, the model $f_{\theta}$ is the mapping from input space to confidence scores. The prediction of the model for input $x$ is the one that provides a maximum confidence score 
\begin{equation}\label{eq-label-ano}
\hat{y}_j = \arg \max_j f_\theta(x)_j
\end{equation}
 where $j$ denotes the $j$-th element of vector $f_\theta(x)$.
 
 \subsection{Pseudo Labeling}
 
 Pseudo labeling is a mechanism to assign labels to unlabeled examples such that artificial labels are further used for training. In the typical pseudo labeling process using consistency regularization, two types of augmentations are used, weak and strong augmentations, denoted by $\alpha(.)$ and $\mathcal{A}(.)$, respectively. Similar to \citet{berthelot2019remixmatch,sohn2020fixmatch} we apply augmentation anchoring  \citet{berthelot2019remixmatch} where pseudo labels are obtained from weakly augmented samples and use the output as an anchor to align the output from a strongly augmented sample to the anchor. During the training, the model is used to predict pseudo labels from strongly augmented unannotated images. The pseudo labels with high confidence for unlabeled inputs are used to train the model. In this approach, weak augmentation of the input images could be as simple as random crop and random horizontal flip, while strong augmentations might include image augmentations methods such as random affine and color jitter with variable intensities. We employ the following loss term
\begin{equation}
\begin{split}
\mathcal{L}(&\mathcal{A}(X_U), \hat{Y}_U; \theta) \\
&:= \sum_{i=L+1}^n \mathbb{I}_{(\max(f_{\theta}(x_i))> \tau_c)} l_s(f_\theta({\mathcal A}(x_i)), \hat{y}_i)
\end{split}
\end{equation}
where we let $\hat{Y}$ denote the set of pseudo labels and $l_s$ is any supervised loss such as cross-entropy loss or MSE loss in the case of soft labels $\hat{Y}_L$. We let $\tau_c$ denote the threshold for the confidence of model predictions and the loss is calculated over the sample outputs with confidence higher than $\tau_c$. In contrast to the approach of \citet{iscen2019label} where the model is firstly trained on annotated data and afterward the labels are assigned according to \eqref{eq-label-ano}, our proposed approach predicts the pseudo labels during the training process. To generate the pseudo labels we use the average of model prediction on $K$ weakly augmented versions of the input sample. In this way, the pseudo label guessing is a more confident process that results in a more stable training process. More technically, for each unlabeled sample $x_u$, the model output produces a pseudo label using equation \eqref{eq-label-ano}. We use the exponential moving average over a sequence of model updates to compute the predictions. 

 \begin{figure*}
    \centering
    \includegraphics[scale=0.3]{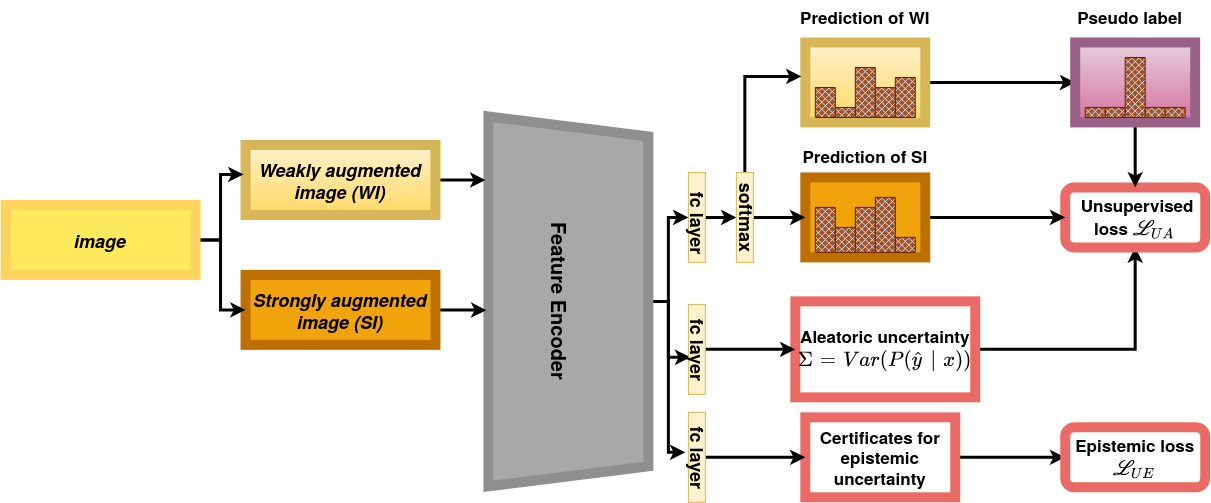}
    \caption{The proposed method for optimizing the model parameters over unlabeled data. The aleatoric uncertainty-aware loss $\mathcal{L}_{UA}$ which minimizes the negative log of Gaussian likelihood loss over pseudo labels generated from weakly augmented data; the epistemic loss  $\mathcal{L}_{UE}$ which maps the projection of certificates generated by the classification model to zero.}
    \label{fig-overall_arch}
\end{figure*}

\section{Method}
We develop an uncertainty-aware loss function that concentrates on the uncertainty from pseudo labels, the uncertainty of the out of distribution unannotated examples, and the uncertainty of the low-density regions due to the inclusion of unlabeled data in the training set. In the following, we demonstrate the proposed uncertainty-aware SSL and describe the major elements of our construction. 

\subsection{Loss}

Our loss function $\mathcal{L}$ has two components corresponding to supervised loss and unsupervised loss. We use the supervised loss $\mathcal{L}_S$ for labeled data and the unsupervised loss $\mathcal{L}_{U}$ for unlabelled data 
\begin{equation}
\begin{split}\label{eq-total-loss}
    \mathcal{L} & = \mathcal{L}_S + \mathcal{L}_{U} \\
    & = \mathcal{L}_S + \alpha_{UA}\,\mathcal{L}_{UA} + \alpha_{UE}\,\mathcal{L}_{UE}
\end{split}
\end{equation}
where $\mathcal{L}_{U} = \alpha_{UA}\,\mathcal{L}_{UA} + \alpha_{UE}\,\mathcal{L}_{UE}$ is the breakdown of unsupervised loss, accounting for the aleatoric and epistemic uncertainty, respectively. $\alpha_{UA}$ and $\alpha_{UE}$ are fixed hyperparameters denoting the weights of the unsupervised losses. In supervised learning, the model is trained by a supervised loss over the annotated data. Our supervised loss $\mathcal{L}_S$ employs the labeled data to minimize the loss and is in the form
\begin{equation}
\mathcal{L}_S(X_L, Y_L; \theta) := \sum_{i=1}^ll_s(f_\theta(x_i), y_i)
\end{equation}
A typical choice for $l_s$ is cross-entropy loss. The unsupervised loss $\mathcal{L}_{U}$ is composed of two parts $\mathcal{L}_{UA}$ and $\mathcal{L}_{UE}$ taking into account the aleatoric uncertainty and epistemic uncertainty, respectively. The unsupervised loss $\mathcal{L}_{UA}$ which is aware of aleatoric uncertainty is the Gaussian likelihood loss with a nontrivial variance to quantify the aleatoric uncertainty. Nevertheless, the epistemic uncertainty loss $\mathcal{L}_{UE}$ measures the uncertainty of the model due to the low-density regions in the feature space and out of distribution unlabeled examples. The output of our model consists of three components as shown in Figure \ref{fig-overall_arch}: the prediction probabilities, the uncertainty of the predictions which predict the diagonal elements of aleatoric uncertainty, and finally the certificate for measuring the epistemic uncertainty. For predicting the two latter components we extend the model after the feature extractor using fully-connected blocks. 

Incorporating the uncertainties in SSL allow the hypothesis over the labeled and unlabeled domains to extend its limit by efficiently evolving the model distribution when observing new data and incorporating the uncertainty from the unlabeled domain and generated pseudo labels. We decouple the aleatoric and epistemic uncertainties in the model predictions and treat each uncertainty with a different loss term. We do not compute aleatoric uncertainty over the annotated dataset with artificial labels, however, the certificates for epistemic uncertainty are trained over the entire dataset.

\subsection{Aleatoric Uncertainty-aware Loss}
We start by incorporating the aleatoric uncertainty to the unsupervised loss function, i.e., the uncertainty related to the conditional distribution of pseudo labels conditioned on the input unlabeled feature sample. To estimate the aleatoric uncertainty we estimate the covariance of uncertainty estimation of the prediction. We train a deep uncertainty covariance matrix using a multivariate Gaussian density loss function. The probability of a pseudo label $\hat{y}$ given the model input $x$ can be estimated with the multivariate Gaussian distribution 
\begin{equation}
\begin{split}
p&(\hat{y}|x) = \frac{1}{\sqrt{(2\pi)^h |\hat{\Sigma}(x)|}} \times \\
&\exp\left({-\frac{1}{2}(\hat{y} - f_{\theta}(x))^T\hat{\Sigma}(x)^{-1}(\hat{y} - f_{\theta}(x))}\right)
\end{split}
\end{equation}
where $f$ is the deep model which provides the mean of distribution, i.e., $E[\hat{y}|x]$. In this formulation, $\hat{y}$ is the pseudo label for the unlabeled sample $x$. The covariance matrix parameters in the log-likelihood loss describe the distribution of model predictions and pseudo labels of unlabeled data with respect to the model prediction. To compute the pseudo labels we follow \citet{sohn2020fixmatch} and we first compute the prediction output using weakly augmented versions of the input sample and then we use the output as a pseudo label for the strongly augmented version of the same sample.

In contrast to the traditional approaches in SSL which only use predictive meanwhile neglecting the predictive variance, we accounted for the variance of prediction with the log-likelihood Gaussian loss. To learn $f$ and $\Sigma$, the optimal parameters minimize the negative log-likelihood equation. The Maximum Likelihood loss is defined as the negative of log-likelihood of predictions using the predicted covariance of Gaussian distribution
\begin{equation}
\begin{split}
& \mathcal{L}_{UA}(X_U, \hat{Y}_U; \theta)  = \\ 
& \frac{1}{n - L}\sum_{i=L+1}^n\mathbb{I}_{(\max(f_{\theta}(x_i))> \tau_c)}\\
& \left(\frac{1}{2}(\hat{y}_i - f_{\theta}(x_i))^T\hat{\Sigma}_i^{-1}(\hat{y_i} - f_{\theta}(x_i)) + \frac{1}{2} \ln |\hat{\Sigma}_i|\right)
\end{split}
\end{equation}
where $\{\hat{\Sigma}_i\}_{L+1 \leq i \leq n}$ which are $h\times h$ covariance matrices for $i-th$ sample, corresponding to the network uncertainty for the output vector $f_{\theta}(x_i)$ and the pseudo label $\hat{y}_i$. We let $\tau_c$ denote the confidence threshold for the model prediction. Therefore we only incorporate the pseudo labels with the confidence more than the threshold $\tau_c$. $\hat{\Sigma}_i$ has $\frac{h(h+1)}{2}$ degrees of freedom due to the symmetry of covariance matrix. We assume a diagonal form for the covariance matrix $\Sigma_i$ across all the samples. I.e., we consider the covariance matrix as the following
\begin{equation}
\hat{\Sigma}_i = diag(e^{2u_{1i}}, e^{2u_{2i}}, \ldots, e^{2u_{hi}}), 
\end{equation}
where we use the Sigmoid function in the model output of covariance to ensure the elements of the vector $({u_{1i}}, {u_{2i}}, \ldots, {u_{hi}})$ are between 0 and 1. The diagonal assumption for the covariance matrix decouples the prediction for each class and neglects the correlation across the class predictions. Although the prediction across classes can be highly correlated we simplify the covariance matrix by neglecting the full multivariate uncertainties which account for cross variances. Regressing the covariance matrix in exponential form removes the singularity around zero in the loss function. In fact, our model learns to regress the uncertainty of the predictions, which further are used to calibrate the MSE loss for unsupervised data during the training. We could also enhance confident learning by accounting for the variance of the predictions using a threshold parameter that could be the focus of future studies. 

\subsection{Epistemic Uncertainty Estimation Loss}
As we described earlier we estimate epistemic uncertainty independent from aleatoric uncertainty. Epistemic uncertainty which is referred also as model uncertainty reflects the uncertainty in the model parameters being mainly due to the low-density regions. Hence, the epistemic uncertainty is reduced by increasing the data and model parameters in the training process. Epistemic uncertainty can be large when the model tries to extrapolate its knowledge on a domain that is significantly different from the data which has been previously presented to the model. For instance, in SSL the strongly augmented unlabeled samples which are used in consistency regularization are significantly different from the weakly augmented examples and therefore the model provides high epistemic uncertainty for these samples. The heavily augmented images are potentially out of distribution, which is in fact beneficial in boosting the performance of SSL \citet{dai2017good}. Nevertheless, in \citet{sohn2020fixmatch} it is found that distribution alignment has a critical impact on achieving state-of-the-art accuracy for datasets such as CIFAR-100. The hypothesis of the model over the distribution of labeled and unlabeled domains can dynamically change when heavily augmented examples are presented to the model. Our ultimate target is to mainly take account of the epistemic uncertainty and minimize the uncertainty when the strongly augmented unlabeled data are fed to the model. 
 In this way, we align the epistemic uncertainty distribution of labeled and unlabeled data and make the model extrapolate its knowledge to the unlabeled data.

To estimate epistemic uncertainty we follow \citet{tagasovska2019single} and leverage orthogonal certificates (OCs) which constitute  a set of diverse functions that project samples to zero. OCs map out of distribution samples to non-zero values, therefore forcing the output of OCs to be zero during the training aligns the distributions of labeled and unlabeled domains. To generate the orthogonal certificates, we consider the deep neural model $y = f_{\theta}(\phi_{\theta}(x))$ where $\phi_{\theta}$ is a feature extractor from the input data $x$ and $f_{\theta}$ is the neural model clustering the features in the embedding space. Given the $n$ samples, we construct the $k$ certificates $C = (C_1, C_2, \ldots, C_k)$ on the $d$ dimensional features of $\phi_{\theta}(x)$ of $n$ training samples, i.e., $\Phi_{\theta} = \{\phi_{\theta}(x_i)\}_{i=1}^n$. Each certificate $C_i$ is a shallow neural model which is a fully-connected layer in our work mapping the features to zero by minimizing the Mean Square Error (MSE) loss. If we use a fully-connected layer for certificates, the certificate is $d\times k$ dimensional layer to predict the zero vector of dimension $k$ for examples drawn from the data distribution. To impose diversity across the certificates, the weights of certificates are trained to be orthogonal. Therefore, the loss function for training the certificate over labeled and unlabeled data is formulated by
\begin{equation}\label{eq-epis-loss}
\begin{split}
\mathcal{L}_{UE}(X_U; \theta) = & \frac{1}{n} \sum_{i=1}^{n} l_c(C^T\phi_{\theta}(x_i), 0)\\
& + \lambda || CC^T - I_k||.
\end{split}
\end{equation} 
Letting $l_c$ be the MSE loss encourages the certificates to be identical to the null space of the feature vector of the samples. As indicated in \citet{tagasovska2019single} the certificates are able to distinguish the out of distribution feature samples $\phi_{\theta}(x)$ which are drawn from Gaussian distribution with the covariance matrix of small associated eigenvalues or small $\norm{CV}$ where $V$ is an eigenvector of the covariance matrix. In SSL the high value of epistemic uncertainty shows that the model is highly confused about the label of unannotated example $x_u$. By incorporating epistemic uncertainty into the loss function, we minimize the discrimination for distribution of unlabeled data with artificial labels when exposing unlabeled data to the neural model. This means that the certificates will be able to discriminate the unlabeled data which has a different distribution than the seen data and minimize the margin among the distributions leading the model to better extrapolate its knowledge over the low-density regions during the training. We use a single neural model that outputs the mean, $\Sigma$ for aleatoric uncertainty-aware loss, and certificate $C$ for computing epistemic uncertainty. 
\begin{table*}[ht]
\centering
\resizebox{0.92\linewidth}{!}{%
\begin{tabular}{lccccc}
\hline
\multirow{2}{*}{Dataset} & \multicolumn{2}{c}{CIFAR-10} & \multicolumn{2}{c}{SVHN} & {CIFAR-100}   \\
                         & 1000 labels  & 4000 labels & 1000 labels  & 4000 labels & 10000 labels    \\
\hline
MeanTeacher \citet{berthelot2019remixmatch} & 82.68\% & 89.64\%  &  96.25\% & 96.61\% & --        \\
MixMatch \citet{berthelot2019mixmatch}   & 92.25\%  & 93.76\%   &  96.73\% & 97.11\% & 71.69\%      \\
ReMixMatch \citet{berthelot2019remixmatch} & 94.27\%  & 94.86\%  & 97.17\% & 97.58\% & 76.97\%       \\
FixMatch \citet{sohn2020fixmatch}  & --       & {\bf 95.69\%} & {\bf 97.64\%} & -- & 77.40\%         \\
SimPLE \citet{hu2021simple} & 94.84\% & 94.95\%   & 97.54\% & 97.31\% & 78.11\%        \\
Ours & {\bf 94.92}\%& 95.46\% & 97.50\% & {\bf 97.57\%} & {\bf 79.23\%}    \\
\hline
\end{tabular}
}%
\caption{CIFAR-10 and CIFAR-100 Top-1 Test Accuracy.}\label{table-cifar-10}
\end{table*}
\subsection{Implementation}

In this section, we demonstrate the performance and efficacy of our proposed uncertainty-aware semi-supervised learning on several standard SSL benchmarks.  We describe the datasets and illustrate the SSL setup for our experiments. We also provide the details of our method implementation and the baseline methods. For the experiments, we follow the benchmarks and the evaluation settings in \citet{hu2021simple,iscen2019label}. Our proposed method is the combination of two approaches, aleatoric and epistemic uncertainty quantification for unlabeled data and pseudo labeling combined with consistency regularization. We use the pseudo annotating method following \citet{sohn2020fixmatch} by applying the model average and taking the average of model predictions of weakly augmented versions of the same unlabeled sample for label guessing. Averaging the outputs from weakly augmented samples makes the label guessing more stable. Similar to \cite{berthelot2019mixmatch}, exponential moving average (EMA) is used for averaging the model parameters to predict the model outputs for label guessing. In all our experiments for generating pseudo labels weak augmentations incorporate standard flip-and-shift augmentation and strong augmentations, comprise of RandAugment \citet{cubuk2020randaugment} where we randomly select a transformation from a predefined set of strong augmentations for each sample. 
\subsection{Datasets}
We evaluate our framework for image classification task on three different datasets: CIFAR-10 \citet{krizhevsky2009learning}, SVHN \citet{netzer2011reading}, CIFAR-100 \citet{krizhevsky2009learning} 
and Mini-ImageNet \citet{vinyals2016matching}. Each dataset is considered under the configuration that the dataset is only partially labeled and unlabeled data is used for SSL. We report the accuracy of classifications on the test sets. 

We split the training set into labeled and unlabeled sets by randomly selecting the samples from each class without replacement. We perform experiments with varying amounts of labeled data for CIFAR-10 and SVHN. We evaluate these datasets with 1000 and 4000 labeled images which are correspondingly equal to 100 and 400 images per class for CIFAR-10. For CIFAR-100 and Mini-ImageNet we use 4000 labeled data. Wide-ResNet and ResNet are used as the backbone architectures in our experiments. Following \citet{berthelot2019mixmatch}, we used a Wide ResNet-28-2 (WRN-28-2) \citet{zagoruyko2016wide} with 1.5M parameters for CIFAR-10 and SVHN, and WRN-28-8 with 23.5M parameters for CIFAR-100. We also use WRN-28-2 and ResNet-18 for Mini-ImageNet. In our implementation, we use the feature extractor combined with a fully connected layer to predict the uncertainty covariance matrix. We also use a linear layer on top of the feature extractor to output the certificates for epistemic uncertainty. We normalize the images for all model inputs to have zero channel-wise mean and unit variance. The performance of deep neural networks trained for classification is heavily determined by the optimizer, scheduler, learning rate, and architecture. These details are not stressed enough in the literature for SSL methods when reporting the empirical evaluations. In our experiments, the initial learning rate is set to $0.01$ for all the datasets. For CIFAR-10, SVHN and CIFAR-100 we use SGD with the initial learning of 0.03 and weight decay of 5e-4 and for Mini-ImageNet we use AdamW \citet{loshchilov2018decoupled} with the initial learning rate of 0.002 and weight decay of 2e-2. For SGD optimization we use cosine scheduler decay with the cosine factor 0.5. We apply the validation set to select the model for the test evaluation. The confidence threshold for the pseudo labeling is set to $0.95$. The weights for the loss function in \eqref{eq-total-loss} use an identical set of hyperparameters $\alpha_{UE} = 1$, and $\lambda = 0.1$ across all the datasets. We set the weight parameters in the loss function $\alpha_{UA} = 75$ for CIFAR-10 and SVHN datasets. We select $\alpha_{UA} = 150$ and $\alpha_{UA} = 300$ for CIFAR-100 and Mini-ImageNet, respectively. The weight selections for $\alpha_{UA}$ is consistent with the weights of unsupervised loss in \citet{hu2021simple}. A key advantage of our method is that it does not require many hyperparameters similar to the previously studied SSL method FixMatch \citet{sohn2020fixmatch} and in particular SimPLE \citet{hu2021simple} which requires setting the threshold parameter for the paired loss across unannotated samples. 
\subsection{Evaluation}
We compare our results with the state-of-the-art results in SSL which include MixMatch \citet{berthelot2019mixmatch}, ReMixMatch \citet{berthelot2019remixmatch}, FixMatch \citet{sohn2020fixmatch}, Label Propagation \citet{iscen2019label} and SimPLE \citet{hu2021simple}. Our method is computationally attractive compared to SimPLE and ReMixMatch since it does not use the paired loss and distribution alignment, while our results are comparable and on par with these approaches. We present a concise comparison of the SSL algorithms in Table \ref{table-cifar-10}. Our results are comparable with the state-of-the-art results from baseline models. Our results for CIFAR-10 and SVHN datasets are on par with the best accuracy from baseline and the result for CIFAR-100 improves the state-of-the-art by $1.12$. We improved the state-of-the-art accuracy for CIFAR-100 while we did not use the distribution alignment term which is found in \citet{sohn2020fixmatch} to be critical to obtaining the best accuracy.
\begin{table}[t]
\centering
\resizebox{0.9\linewidth}{!}{%
\begin{tabular}{lccccc}
\hline
\multirow{2}{*}{Dataset} & \multicolumn{2}{c}{Mini-ImageNet}  \\
                         & 4000 labels & architecture  \\
\hline
MixMatch & 55.47\%  & WRN 28-2      \\
SimPLE   & 66.55\%    &  WRN 28-2   \\
Ours & {\bf 66.81\%}      & WRN 28-2       \\
\hline
MeanTeacher  & 27.49\%  & ResNet-18  \\
Label Propagation  & 29.71\% & ResNet-18      \\
SimPLE & 49.39\%   & ResNet-18    \\
Ours & {\bf 50.75\%}     & ResNet-18     \\
\hline
\end{tabular}
}%
\caption{Mini-ImageNet Top-1 Test Accuracy.}\label{table-mini-imagenet}
\end{table}
We examine the performance and the scalability of our uncertainty-aware SSL for Mini-ImageNet in Table \ref{table-mini-imagenet} where we listed the performance of our method along the baselines for WRN 28-2 and ResNet18 architectures. We also reported the results from the baseline \citet{iscen2019label} for a fair comparison. Our proposed approach outperforms other approaches by a large margin which shows the efficiency of capturing uncertainties due to the acquisition of the unlabeled domain for gaining performance. Our method outperforms SimPLE despite being simpler due to removing the paired loss which is the main contribution of SimPLE. 
\subsection{Ablation Study}
In the following, we perform an ablation study on the proposed method over the CIFAR-10 and CIFAR-100 datasets with WRN 28-2 and WRN 28-8 backbones, respectively. Our aim is to study the effect of different components of loss function \eqref{eq-total-loss} in performance improvements. In Table \ref{table-ablation-study} we listed the results from different loss components. The results show that our uncertainty-aware method outperforms the baseline method (without uncertainty losses) by a significant margin. The reason is that we incorporate the uncertainty from the unlabeled domain directly to the loss function, therefore the SSL is robust to the inherited noise and uncertainty from pseudo labels and out of distribution unlabeled samples. It is observed from Table \ref{table-ablation-study} that the aleatoric uncertainty loss has a significant impact on accuracy. This result pronounces that the performance gain is mainly due to accounting for the aleatoric uncertainty when the model's knowledge is extrapolated from labeled samples through noisy pseudo labels to the unlabeled datasets. It also shows that the performance enhancements using a higher weight for the orthogonality regularization in the epistemic loss \eqref{eq-epis-loss} which naturally encourages the diversity across the certificates. 
\begin{table}[t]
\centering
\resizebox{0.95\linewidth}{!}{%
\begin{tabular}{lcccc}
\hline
\multirow{2}{*}{Dataset} & \multicolumn{2}{c}{CIFAR-10}   & \multicolumn{2}{c}{CIFAR-100}   \\
                         & 4000 labels & architecture & 10000 labels &  architecture   \\
\hline
Ours &  95.46\% & WRN 28-2 & 79.23\%   & WRN 28-8\\
w/o $\mathcal{L}_{UE}$   & 95.23\%  &  WRN 28-2  & 75.36\%  & WRN 28-8       \\
w/o $\mathcal{L}_{UA}$ & 94.90\% & WRN 28-2 & 73.06\% & WRN 28-8       \\
{w/o $\mathcal{L}_{UE}$ -- w/o $\mathcal{L}_{UA}$} & {94.97\%}  & {WRN 28-2} & {72.72\%}  & {WRN 28-8}\\\hline
 $\lambda = 0.5$ & 95.99 & WRN 28-2 & 79.44 & WRN 28-8\\
\hline
\end{tabular}
}%
\caption{Ablation study on CIFAR-10 and CIFAR-100.}\label{table-ablation-study}
\end{table}
\section{Conclusion}
We have proposed an approach that relies on aleatoric and epistemic uncertainty quantification to account for uncertainties of unlabeled data when the network's knowledge is extrapolated to  the unlabeled domain. We show that the proposed uncertainty-aware objective could contribute to gaining performance and improve the state-of-the-art methods for semi-supervised learning. Our method outperforms the results from state-of-the-art SSL baseline models on Mini-ImageNet, CIFAR-100, and CIFAR-10 datasets and is on par with the recent results for SVHN while being computationally more efficient. Our method can be applied as complementary to SSL methods to further improve the performance. 

%%%%%%%%% REFERENCES
{\small
\bibliography{Refs}
}

\end{document}